\documentclass[letterpaper, 10 pt, conference]{ieeeconf}
\IEEEoverridecommandlockouts
\usepackage{amsmath,amsfonts}
\usepackage[ruled]{algorithm2e}
\usepackage{array}
\usepackage{textcomp}
\usepackage{stfloats}
\usepackage{url}
\usepackage{verbatim}
\usepackage{graphicx}
\usepackage{cite}
\providecommand{\FIG}[1]{Fig.~\ref{#1}}
\providecommand{\SEC}[1]{Section~\ref{#1}}

\newcommand{\ie}{i.e.~}

\usepackage{multirow}
\usepackage{url}
\usepackage{comment}
\usepackage[belowskip=-7pt,aboveskip=3pt]{caption}
\usepackage{subcaption}
\usepackage{booktabs}
\usepackage{tikz}
\usepackage{acronym}
\usepackage{siunitx}

\acrodef{ICP}{iterative closest point}
\acrodef{DOF}{degrees of freedom}
\acrodef{GNSS}{global navigation satellite system}
\acrodef{RTK}{real time kinematics}
\acrodef{UGV}{unmanned ground vehicle}
\acrodef{UAV}{unmanned aerial vehicle}
\newacroindefinite{UAV}{a}{an}
\acrodef{IMU}{inertial measurement unit}
\acrodef{SLAM}{simultaneous localization and mapping}
\acrodef{APE}{absolute pose error}
\acrodef{RPE}{relative pose error}
\acrodef{FOV}{field of view}
\acrodef{EKF}{extended Kalman filter}
\acrodef{APDGICP}{adaptive probability distribution-GICP}
\acrodef{NDT}{normal distributions transform}
\acrodef{DNN}{deep neural network}
\acrodef{CNN}{convolutional neural network}
\acrodef{GICP}{generalized ICP}

\begin{document}

\bstctlcite{IEEEexample:BSTcontrol} 

\title{Do we need scan-matching in radar odometry?}

\author{Vladimír Kubelka, Emil Fritz and Martin Magnusson
  \thanks{This work was supported by Sweden’s Innovation Agency under grant number  2021-04714 (Radarize).
    The authors would also like to express their gratitude to Annika Nilsson for her part in the implementation and experimental evaluation of this work.}
\thanks{Vladimír Kubelka ({\tt\small{vladimir.kubelka@oru.se}}) and Martin Magnusson ({\tt\small{martin.magnusson@oru.se}}) are with the MRO lab of the AASS research centre at Örebro University, Sweden. Emil Fritz is with Örebro University, Sweden.}%
}

\maketitle

\begin{abstract}
  There is a current increase in the development of ``4D'' Doppler-capable radar and lidar range sensors that produce 3D point clouds where all points also have information about the radial velocity relative to the sensor.
  4D radars in particular are interesting for object perception and navigation in low-visibility conditions (dust, smoke) where lidars and cameras typically fail.
  With the advent of high-resolution Doppler-capable radars comes the possibility of estimating odometry from single point clouds, foregoing the need for scan registration which is error-prone in feature-sparse field environments.
  We compare several odometry estimation methods, from direct integration of Doppler/IMU data and Kalman filter sensor fusion to 3D scan-to-scan and scan-to-map registration, on three datasets with data from two recent 4D radars and two IMUs.
  Surprisingly, our results show that the odometry from Doppler and IMU data alone give similar or better results than 3D point cloud registration. 
  In our experiments, the average position error can be as low as 0.3\% over 1.8 and 4.5$\,$km trajectories.
  That allows accurate estimation of 6DOF ego-motion over long distances also in feature-sparse mine environments.
  These results are useful not least for applications of navigation with resource-constrained robot platforms in feature-sparse and low-visibility conditions such as mining, construction, and search \& rescue operations.
\end{abstract}


\section{Introduction}

Rapid development in millimeter wave imaging radars, driven by the automotive industry, has enabled localization and mapping in environments where we expect deteriorated visibility conditions and dirt deposition on the sensors.
Deploying autonomous vehicles in the mining industry, construction or search \& rescue are example applications that demand such capability.
Modern imaging radars, similarly to 3D lidars, provide 3D scans of the surroundings. 
They are additionally able to estimate the radial velocity of each sensed 3D point by leveraging precise phase measurements of the returning signal. 
This Doppler velocity, as we further denote it, has proven to be advantageous for odometry methods, aiding in the segmentation of dynamic and static objects \cite{kingery2022improving}, as well as introducing more constraints to the ego-motion estimation \cite{DoerMFI2020, Michalczyk2022}.
Moreover, the velocity measurement comes without the need to perform data association, which can be challenging in feature-sparse environments, such as underground mines.

\begin{figure}
    \centering
    \includegraphics[angle=0, width=0.95\linewidth]{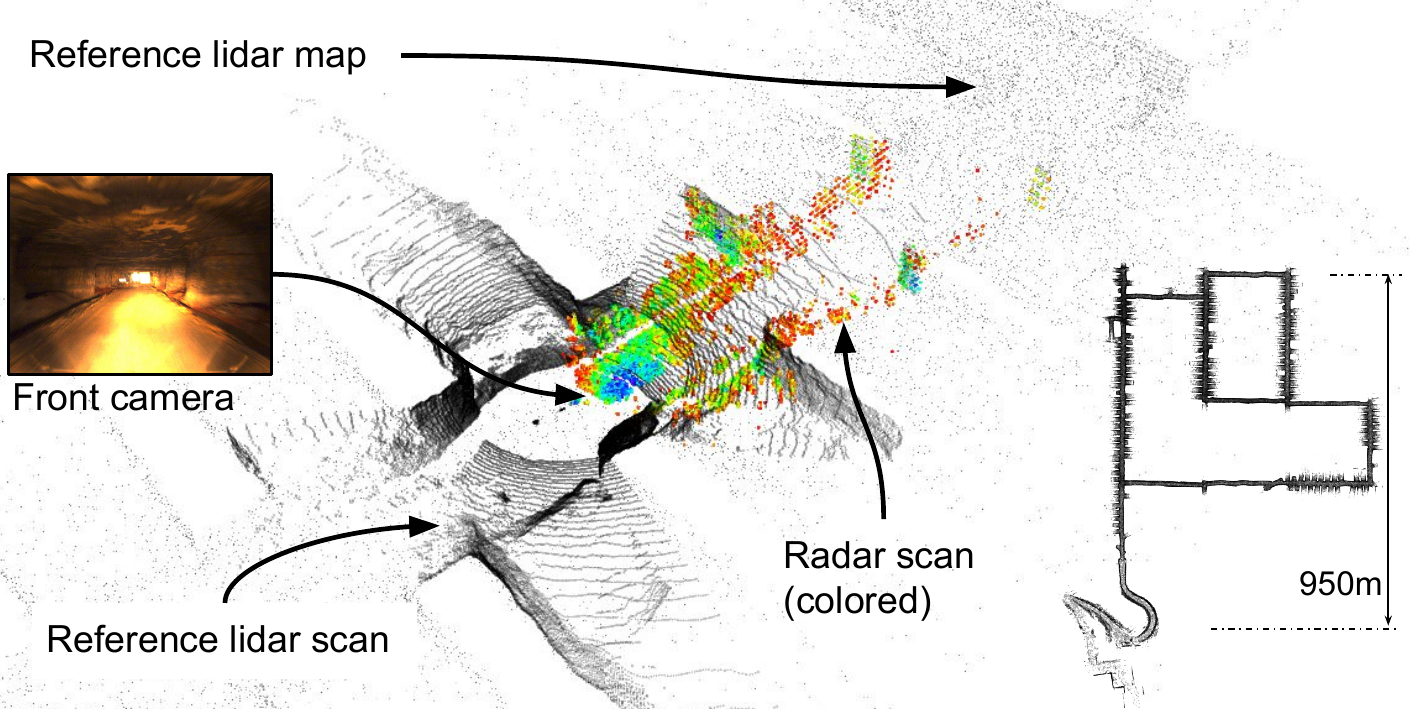}
    \caption{
     Detail from the Kvarntorp  mine environment captured by two sensor modalities, lidar and radar. The radar modality suffers from limited \ac{FOV}, lower resolution and fewer returns. It is however more suitable for low-visibility conditions expected in mining. 
    }
    \label{fig:mine_detail}
\end{figure}

In recent years, several approaches to radar odometry and \ac{SLAM} have emerged.
Motivated by the problem of developing a \ac{SLAM} system for an underground mining environment, we compare several representative radar odometry estimation methods.
To that end, we deploy them on three datasets that include two distinct modern imaging radars.
Two datasets have been recorded with our mobile sensor rig: in an underground mine (\FIG{fig:mine_detail}) and in an outdoor testing site for large wheel loaders (Figs. \ref{fig:volvo_detail} and \ref{fig:volvo_data}).
The third dataset has been published by Zhang et al.~\cite{Zhang2023} and represents a structured urban environment.
Surprisingly, using the simplest method of directly fusing the Doppler-based radar ego-velocity with the orientation provided by an \ac{IMU}, we are able to achieve localization drift as low as 0.3\% over \SI{4.5}{\kilo\meter} and \SI{1.8}{\kilo\meter} trajectories from the mine and the outdoor testing site.  
We find this experimental result useful for designing localization and mapping systems for the mentioned applications and worth spreading in the robotics community for further investigation.
Moreover, we make our dataset publicly available at \url{https://github.com/kubelvla/mine-and-forest-radar-dataset} as the high-grade radars are still difficult to obtain.

\section{RELATED WORK}

In this section, we briefly review radar odometry algorithms based on 2D scanning radars.
Then, more closely relevant to this article, we focus on the state of the art in modern 4D radar odometry.   

Classical 2D scanning radars usually provide the scans in the form of \emph{spectral images}, which encode signal intensity in the radial direction for every azimuth they measure during a single radar scan.
Cen et al.~\cite{Cen2018, Cen2019} proposed methods to extract radar key points from such spectral images and showed how their approach improves the scan matching.
Later works such as Barnes et al.~\cite{Barnes2020} include machine-learning techniques to improve the key point detection.
Burnett at al.~\cite{burnett2021we} shows the importance of addressing motion distortion and Doppler shift in the radar data.
Contrary to searching for key points, Adolfsson et al.~\cite{adolfsson2021cfear} approach the problem from the point cloud perspective, focusing on local geometry that could better constrain the matching process.
Park et al.~\cite{park2020pharao} avoid point matching altogether by applying Fourier-Mellin transform to find correlation between subsequent radar scans.

The rapid development of 4D Doppler-capable imaging radars opens new possibilities in object detection, motion estimation and localization.
The surveys by Venon et al.~\cite{venon2022millimeter} and Zhou et al.~\cite{Zhou2022} provide comprehensive overview of the state of the art in this research direction.
The mmWave sensors by Texas Instruments have gained a lot of attention, as they are lightweight and still provide tens of 3D points with  Doppler velocity.
Doer and Trommer~\cite{DoerMFI2020} propose a loosely-coupled \ac{EKF} filtering method, which fuses radar ego-velocity, inertial and barometric measurements to track the pose of \iac{UAV}.
They develop a RANSAC-based least squares optimization algorithm to extract the radar ego-velocity from the radar data.
In their later work \cite{Doer2022}, they incorporate \ac{GNSS} measurements as well.
In our work, we adopt their open-source implementation to perform experiments with our sensor suite.  
Contrary to a filtering approach, Kramer et al.~\cite{kramer2020radar} propose a sliding window optimization algorithm that fuses Doppler and inertial measurements to track a pose of a mobile sensor rig.
They verify their results in underground and outdoor environments.
Focusing on the \ac{UAV} application, Michalczyk et al.~\cite{Michalczyk2022, Michalczyk2023} propose a tightly-coupled, \ac{EKF}-based radar odometry.
Their approach includes point matching, where the perceieved distance between the matched points, together with the Doppler velocity, serves as the residual vector for the EKF.
Lu et al.~\cite{Lu2020} takes a different direction, both radar measurements and inertial measurement are processed by a \ac{DNN} to estimate pose of a mobile agent.
They use a \ac{CNN} to extract features from the radar data and a recurrent network to analyze the inertial data.
The features are fused in following \ac{DNN} stages to produce pose estimates.

As the 4D imaging radars evolve and their resolution increases, classical 3D scan-matching methods have become feasible.
Zhuang et al.~\cite{Zhuang2023} fuse inertial data, Doppler-based ego-motion estimates and scan-to-submap constraints by an iterative \ac{EKF} to obtain radar odometry.
In a separate module, they complete the system into a full \ac{SLAM} solution by performing a \ac{GICP}-based loop closure and global map optimization.
They use a Continental ARS548 radar with \SI{300}{\meter} range and \SI{0.3}{\meter} resolution that produces approximately 400 points per scan.
Zhang et al.~\cite{Zhang2023} choose a classical \ac{SLAM} approach with their Oculii Eagle radar sensor (approximately 5000 points per scan in their public dataset).
They modify the \ac{SLAM} network from Koide et al.~\cite{Koide2019} by adding a modified \ac{GICP} matching algorithm that takes into account the specific spatial uncertainties in radar point clouds.
Since their work and dataset are open-source, we include them in our experimental evaluation.

\begin{figure}
    \centering
    \includegraphics[angle=0, width=0.95\linewidth, trim=0 90 0 40, clip]{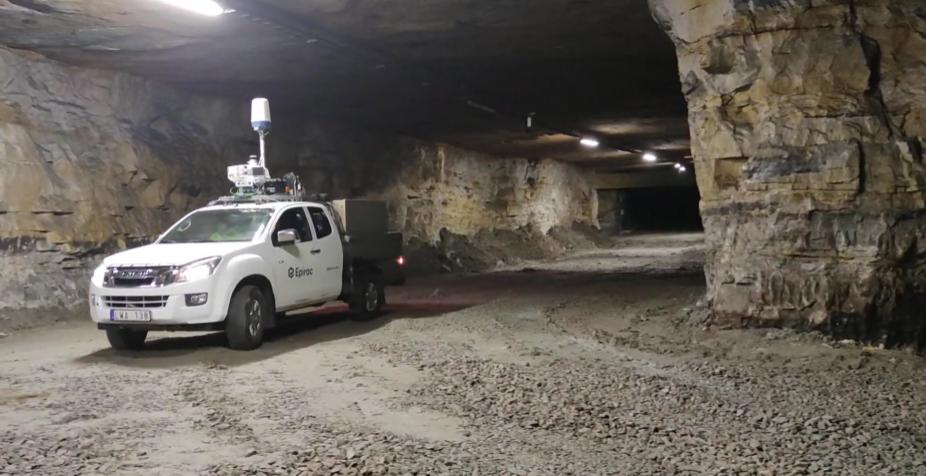}
    
    \vspace{3pt}
    
    \includegraphics[angle=0, width=0.95\linewidth, trim=0 20 0 0, clip]{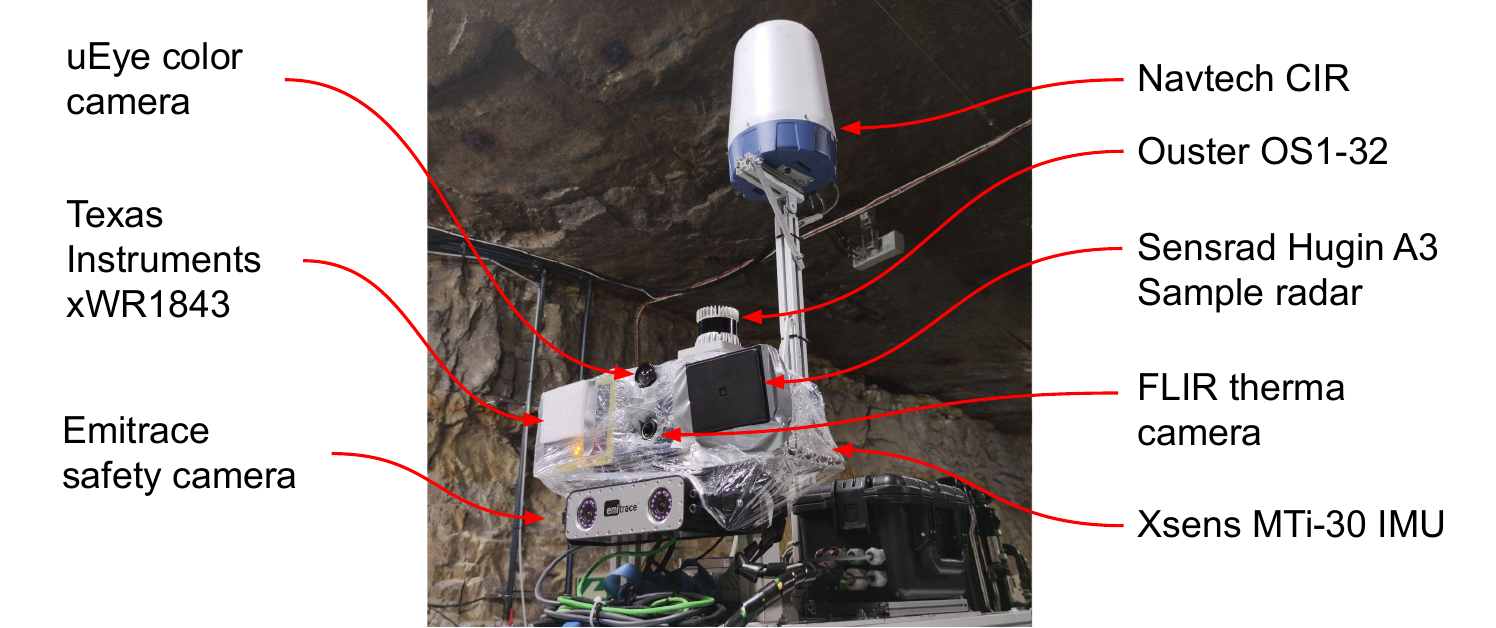}
    \caption{
    The pick-up truck driving through the Mine with the multi-sensor rig attached to the roof (top). The sensor rig detail (bottom).
    }
    \label{fig:sensor_rig}
\end{figure}

\section{Radar odometry variants}
\label{sec:odom_variants}

\begin{figure}
    \centering
    \includegraphics[angle=0, width=0.95\linewidth]{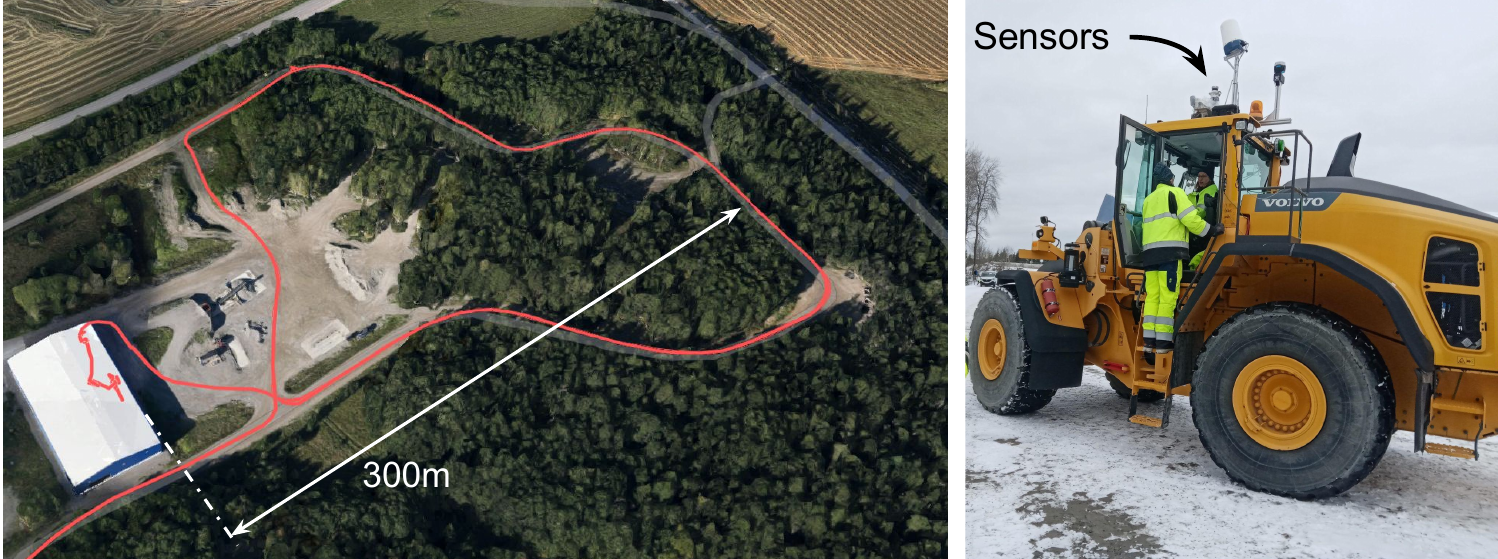}
    \caption{
    Aerial view of the Forest environment (left, source: Google Maps\textsuperscript\textcopyright) and the Volvo CE wheel loader equipped with the sensor rig (right).
    }
    \label{fig:volvo_detail}
\end{figure}

From the variety of radar odometry methods, we choose a representative set which is available open-source, applicable to our sensors and spans from simple sensor fusion to advanced scan-matching.

\subsection{Doppler velocity and IMU}
\label{sec:imu_doppler}

The simplest approach to pose estimation tested in this work exploits the orientation provided by an \ac{IMU} and ego-velocity as measured by a Doppler-capable radar sensor.
The ego-velocity is first transformed from the coordinate frame of the moving platform to the world coordinate frame based on the IMU attitude.
It is then numerically integrated assuming constant velocity between consecutive radar scans.
This way, a trajectory expressed in the world coordinate frame is generated.
Further in the text, we refer to this approach as to \emph{IMU+Doppler}.

Since the radar does not directly provide the ego-velocity measurement but rather radial components of its detected target velocities, it is necessary to robustly process this information to estimate the ego-velocity of the radar.
For this purpose, we deploy the approach and implementation\footnote{\url{https://github.com/christopherdoer/reve}} by Doer and Trommer~\cite{DoerMFI2020}.
Their \emph{3-Point RANSAC-LSQ} ego-motion estimation method applies RANSAC to the underlying least squares optimization problem (eq. 27--32 in \cite{DoerMFI2020}).
This algorithm is highly efficient. The average processing time for one radar scan is \SI{10}{\milli\second} in our dataset.

\begin{figure}
    \centering
    \includegraphics[angle=0, width=0.8\linewidth]{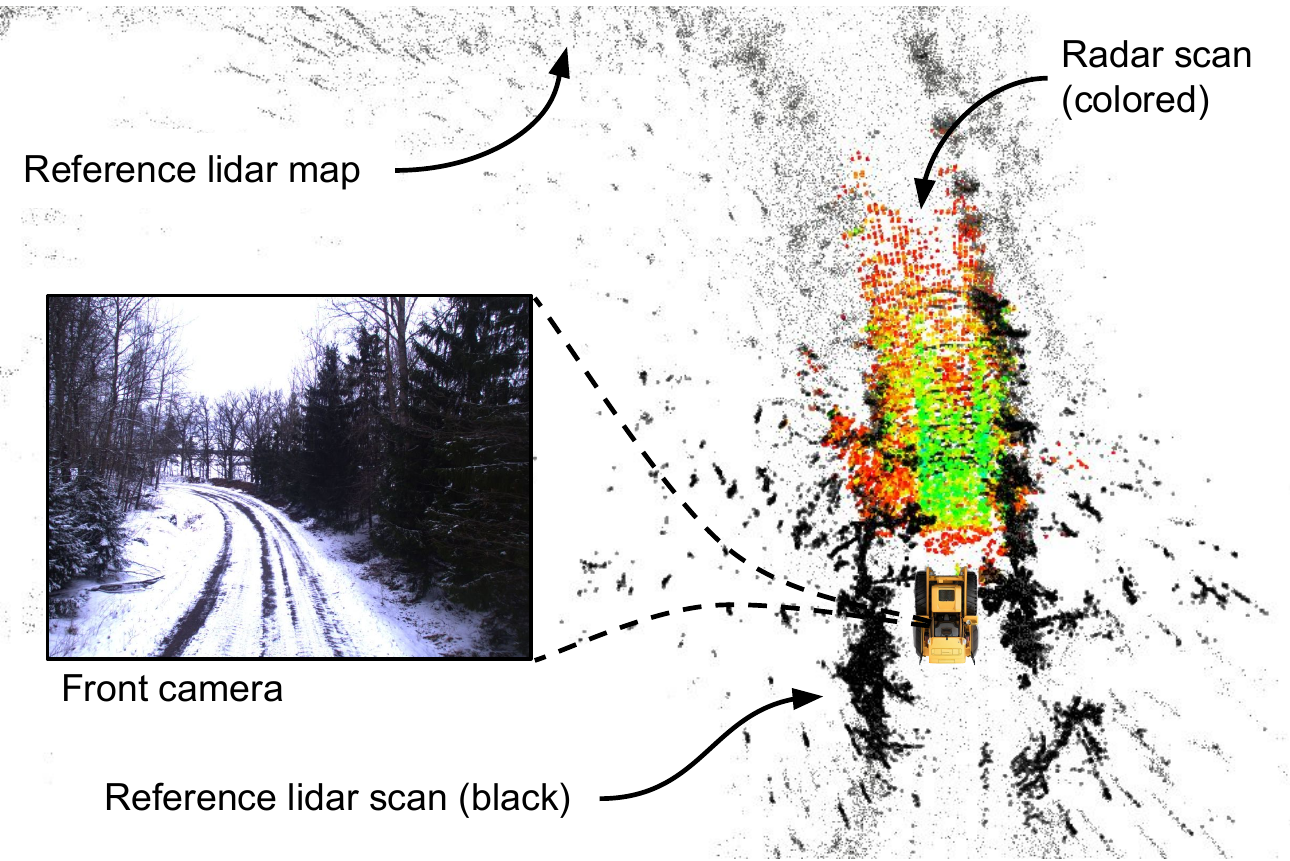}
    \caption{
    Top-view of a yellow wheel loader navigating a forest road. From its suite of sensors, lidar, radar and front camera are shown. The dark black points denote the active lidar scan, which is in contrast with the colored, front-facing radar scan.
    }
    \label{fig:volvo_data}
\end{figure}

\begin{figure}
    \centering
    \includegraphics[angle=0, width=0.9\linewidth, trim=0 0 0 0, clip]{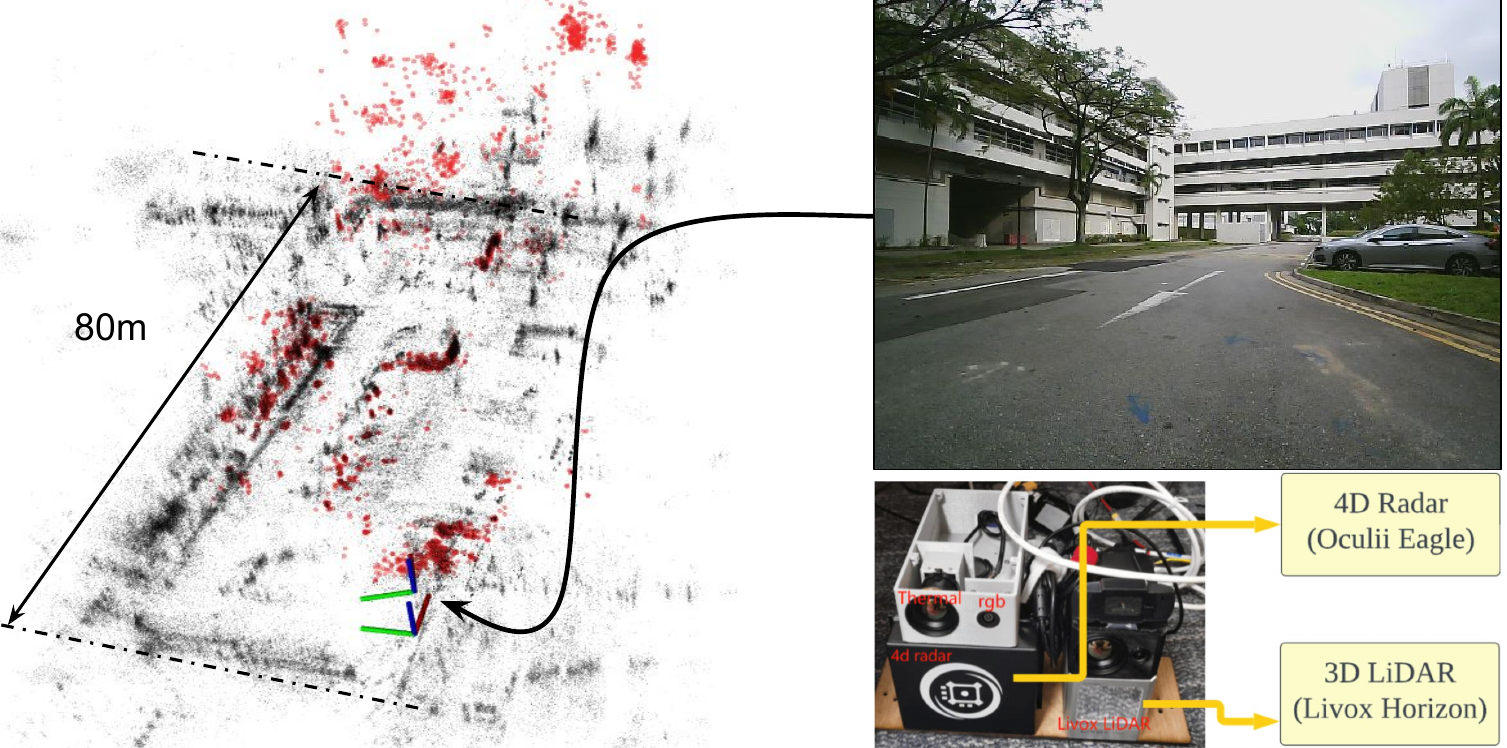}
    \caption{
    Assembled radar map of the Car Park environment (left) denoted by black points with a live radar scan (red points). The sensor rig (bottom right) was manually pushed along the trajectory. The data and figures adopted from \cite{Zhang2023}.
    }
    \label{fig:carpark_detail}
\end{figure}

\subsection{Extended Kalman Filter fusion}
\label{sec:ekf_odom}
In contrast to direct Doppler+IMU fusion,
employing the \ac{EKF} allows more principled handling of noise in the sensor measurements and provides pose confidence estimates.
We benefit from the implementation\footnote{\url{https://github.com/christopherdoer/rio}} by Doer and Trommer~\cite{DoerMFI2020} which combines their 3-Point RANSAC-LSQ ego-motion estimation with inertial and barometric measurements.
The sensor measurements are fused in a loosely-coupled manner by an \ac{EKF}.
There are several extensions of the algorithm available. 
We choose the original \emph{ekf-rio} version as it does not require a precise radar trigger signal, which we unfortunately do not get from our radar.
The algorithm applies, in that case, the incoming ego-motion measurements with a lag of approximately \SI{100}{\milli\second} which can impede the state estimation quality, especially during highly dynamical motion.
Moreover, we omit the barometer measurements as our sensor rig lacks this sensor.
The results we obtain here therefore represent a lower bound on the odometry quality achievable by the filtering methods.
We refer to this approach as to \ac{EKF} in the following text.

It is noteworthy that the works of Michalczyk et al.~\cite{Michalczyk2022, Michalczyk2023} report improvements upon \cite{DoerMFI2020} by employing a tightly-coupled \ac{EKF} filtering for radar-inertial odometry.
They are able to achieve localization drift below 1\%.
It remains an interesting question how the tightly-coupled algorithm handles the high-grade radar scans of thousands of targets.

\subsection{Point-to-plane Iterative Closest Point with local map}
\label{sec:norlab_icp}

The high resolution of the tested radars allow us to test methods originally developed for registration of lidar point clouds.
For testing the scan-to-submap matching variant, we use the \emph{norlab-icp-mapper}\footnote{\url{https://github.com/norlab-ulaval/norlab_icp_mapper}} which is open-source and highly configurable.
It supports a range of \ac{ICP} variants, from which we choose the \emph{point-to-plane} variant as it generally performs well in structured and semi-structured environments.
This mapper does not support map optimization by loop closure identification, it rather builds a monolithic map and thus behaves as a lidar odometry method.
The mapper is set to add new points into the map up to a maximum density defined by the minimum distance between points, which is \SI{0.1}{\meter} in our experiments.
 Point-to-plane \ac{ICP} also requires estimation of normal vectors based on local geometry around each point in the map.
In our experiments, we use 15 nearest points for that.
Also, preliminary tests have shown that this mapper requires a prior motion estimate when deployed on the radar data.
We thus provide the Doppler+IMU pose as the prior in all experiments.

The \ac{ICP} algorithm in the mapper offers full 6-\ac{DOF} pose estimation, or a constrained 4-\ac{DOF} pose estimation.
In the 4-\ac{DOF} variant, only position and heading are optimized in the point cloud registration, the other two \ac{DOF} are directly adopted by the mapper from the \ac{IMU}-provided orientation.
In this work, we test both variants, and refer to them as to \emph{ICP} and \emph{ICP 4DOF}.

\subsection{Scan-to-scan matching variants}
\label{sec:gicp_variants}

The final group of radar odometry variants tested in this work employs the scan-to-scan matching, which is often used in front-end modules of larger \ac{SLAM} frameworks.
Zhang et al.~\cite{Zhang2023} successfully applies this approach in a \ac{SLAM} framework with a modern imaging radar (Oculii Eagle).
Since their implementation of the \ac{SLAM} framework is open-source\footnote{\url{https://github.com/zhuge2333/4DRadarSLAM}}, we include it here for testing their radar odometry with our radar dataset.
Moreover, they provide one session from their dataset which in turn allows us to test all the other approaches with the Oculii Eagle radar.
Their radar odometry front-end is highly configurable, allowing  users to choose from several other scan-matching algorithms.
We choose to test their \ac{APDGICP} variant of \ac{GICP}.
Their scan-matching method can function without a prior motion estimate, yet we modify the code to include the option to use the Doppler+IMU odometry prior.
This makes the comparison with the scan-to-submap-matching variants fair.
When providing the prior, we refer to the method as to \emph{APDGICP IMU Prior}, \emph{APDGICP} otherwise.

We also choose to test their implementation \ac{NDT} scan-matching algorithm as it is often used in lidar odometry solutions.
For \ac{NDT}, we always use the Doppler+IMU prior and refer to it as to \emph{NDT} in the evaluation.

\section{Experiments and Analysis}
\label{sec:experiments}

\begin{figure}
    \centering
    \includegraphics[width=0.49\textwidth]{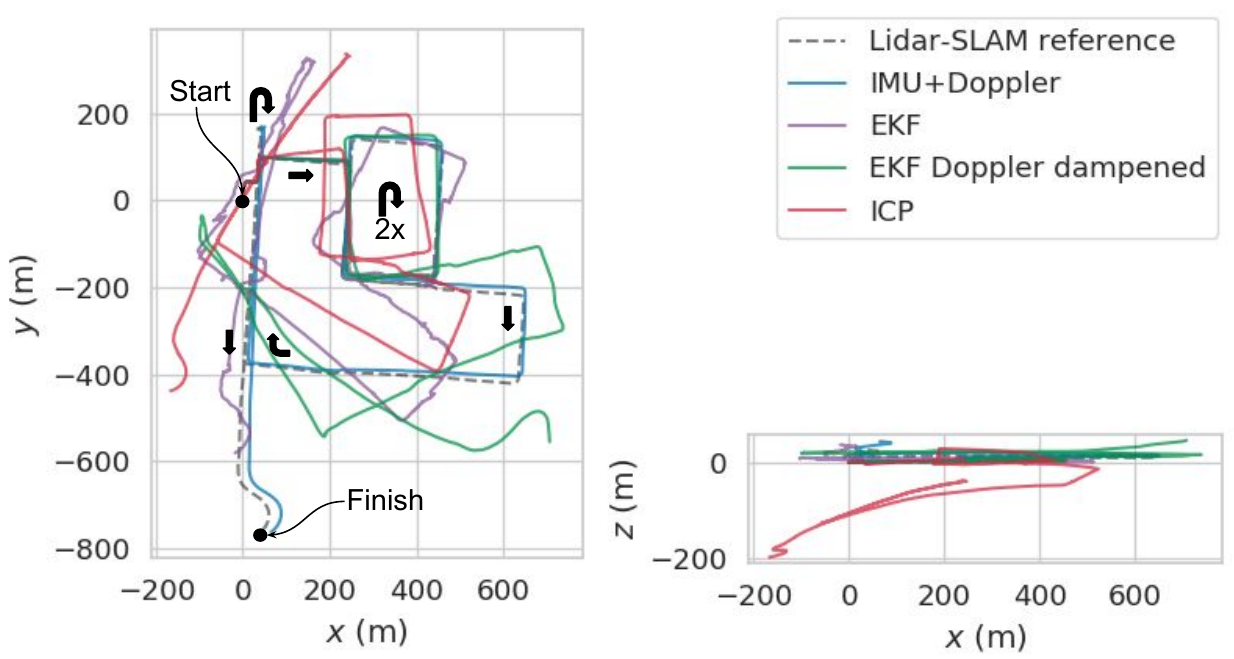}
    \caption{The Mine environment recorded with the Sensrad Hugin radar. The 4DOF ICP is omitted from this plot for clarity, its vertical drift would be limited compared to the standard ICP shown in red. Similarly, the scan-to-scan matching odometries are not shown for their fast divergence.}
    \label{fig:trajectory_mine}
\end{figure}

\begin{figure}
    \centering
    \includegraphics[width=0.45\textwidth]{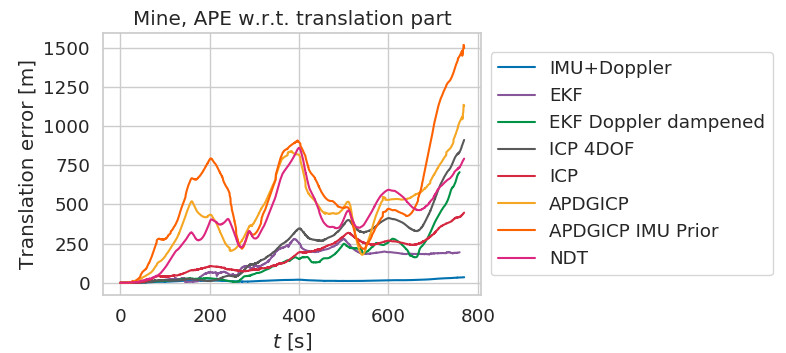}
    \caption{The translation component of APE in the Mine for all discussed odometry variants.}
    \label{fig:ape_kvarn}
\end{figure}

\begin{figure}
    \centering
    \includegraphics[width=0.45\textwidth]{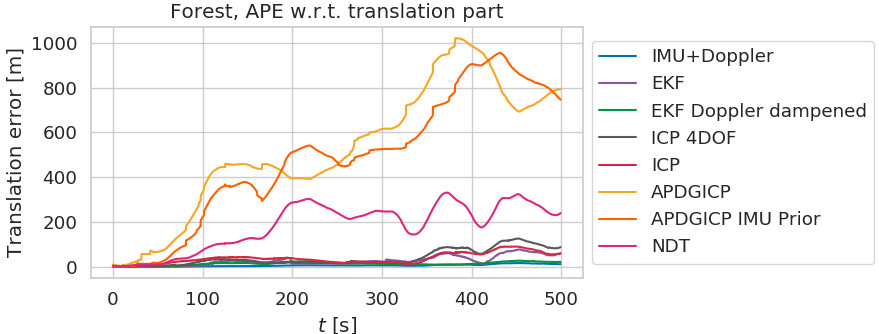}
    \caption{The translation component of APE in the Forest for all discussed odometry variants.}
    \label{fig:ape_forest}
\end{figure}

\begin{figure}
    \centering
    \includegraphics[width=0.49\textwidth]{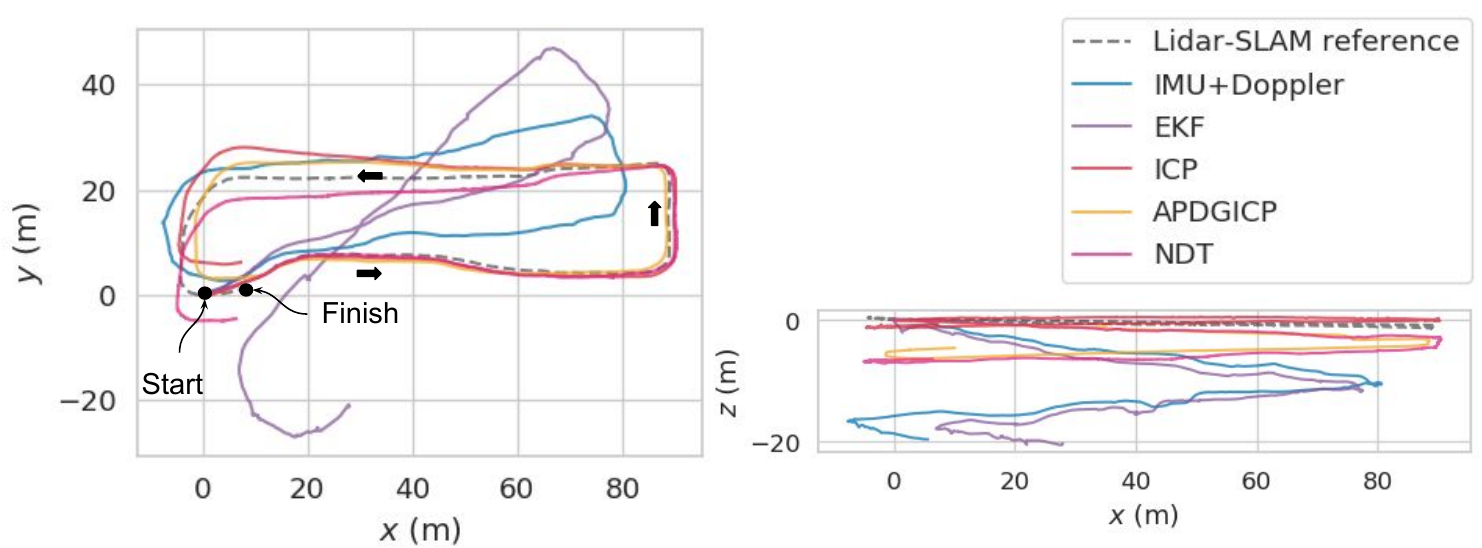}
    \caption{The Car Park experiment from \cite{Zhang2023} recorded with the Oculii Eagle radar. Only selected odometry variants shown for better clarity.}
    \label{fig:trajectory_cp}
\end{figure}

\begin{figure}
    \centering
    \includegraphics[width=0.45\textwidth]{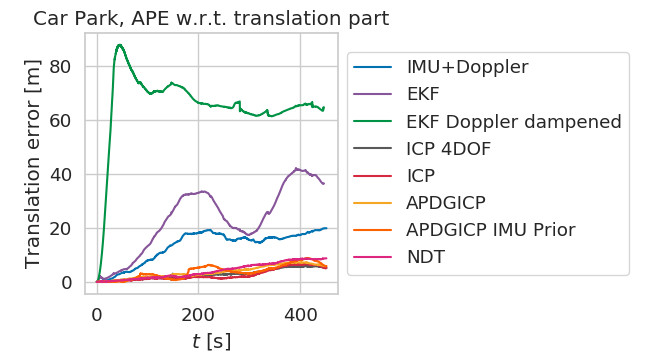}
    \caption{The APE values for the Car Park environment for all discussed odometry variants.}
    \label{fig:ape_cp}
\end{figure}

This section first introduces the two environments used for recording the experimental trajectories and then one experiment adopted from the dataset published by Zhang~\cite{Zhang2023}. The details about the sensor suites used to record the experiments are provided as well.
Subsequently, the performance of the discussed radar odometry approaches is studied by the means of the \ac{APE} and \ac{RPE} metrics which are widely used for this purpose.

\subsection{Environment and sensor setup}
\label{sec:env_sens}

Motivated by research towards \ac{SLAM} in harsh environmental conditions, two field datasets were recorded: one in the Kvarntorp research mine outside of Örebro, Sweden, and one at an outdoor testing site for Volvo Construction Equipment wheel loaders and dumpers in Eskilstuna, Sweden.

\textbf{The Kvarntorp test mine} provides a model environment for underground mining industry applications.
A \SI{4500}{\metre}-long run was recorded with a sensor rig attached to the roof of a pick-up truck as shown in \FIG{fig:sensor_rig}.
The average speed was \SI{21}{\kilo\metre\per\hour} which is close to the max safely drivable speed in the mine.
\FIG{fig:mine_detail} gives a general impression of the underground tunnels.
At some locations, the tunnels are straight with no side-tunnels and these sections are generally the most demanding for any kind of \ac{SLAM}, regardless the modality.
On the other hand, locations with side-tunnels provide a large number of geometrical constraints a \ac{SLAM} algorithm can benefit from.
\FIG{fig:mine_detail} shows such an area, with two modalities presented, lidar in greyscale points and radar in colored points.
For reference, the view from an RGB camera is also provided.
Further in the text, we will refer to this experiment as \emph{Mine}.

\textbf{The Eskilstuna outdoor testing site} is used by Volvo CE for development and testing of their products, including the large wheel loaders as shown in \FIG{fig:volvo_detail}.
For our experiments, the sensor rig used in Mine was moved from the pickup truck to the Volvo wheel loader.
A \SI{1800}{\metre}-long trajectory was recorded which took the wheel loader through open space and on a forest road (see \FIG{fig:volvo_data}).
The trajectory was a loop, repeated twice, with the average speed of \SI{13.6}{\kilo\metre\per\hour}.
Precise RTK-GPS reference was recorded but for the purposes of this study, we base our metrics on a lidar-\ac{SLAM}-based reference, which provides the full 6-\ac{DOF} pose.
We only confirm here that the positioning from the RTK-GPS agrees with our reference SLAM results.
Further on in the text, we will refer to this testing site as to \emph{Forest}.

\textbf{The sensor suite} used in the experiments is detailed in \FIG{fig:sensor_rig}.
The sensors are attached to a metal rig and connected to an Intel NUC computer that runs Ubuntu with ROS installed.
Raw data are saved to ROS bag files for later processing.
The suite consists of three radars, one lidar, three cameras and an \ac{IMU}.
The radar used in Mine and Forest is the Sensrad Hugin A3 radar, with horizontal and vertical \ac{FOV} \SI{80}{\degree} and \SI{30}{\degree} respectively.
Thanks to its configuration of 48 $\times$ 48 transmitting and receiving antennas, the horizontal and vertical resolution is \SI{1.25}{\degree} and \SI{1.7}{\degree}.
The radar is operated in \emph{short range} settings, which implies maximum range \SI{42}{\meter}, but grants the highest range resolution of \SI{0.1}{\meter}.
The frame rate is \SI{16}{\hertz} and the scans contain approximately 10000 points in our environments.
For reference localization, an Ouster OS1-32 lidar is used. The lidar frame rate is \SI{10}{\hertz} and the all point are time-stamped under PTP synchronization with the master computer.
Finally, inertial data are recorded by an Xsens MTi-30 \ac{IMU} at \SI{400}{\hertz} rate.
The \ac{IMU} is running its own attitude estimation using the \emph{VRU General} profile, which does not use magnetometer data to absolutely reference the heading angle.
Yet, the magnetometer measurements are still used to estimate gyro biases and thus limit the heading drift down to \SI{3}{\degree\per\hour} in ideal conditions.

\textbf{The Car Park} trajectory is a part of the dataset recorded by Zhang et al.~\cite{Zhang2023}.
In their setup, they used the Oculii Eagle radar that provides \ac{FOV} of \SI{120}{\degree}$\times$\SI{30}{\degree} (horizontal, vertical) and resolution of \SI{0.5}{\degree}, \SI{1}{\degree} and \SI{0.16}{\meter} (horizontal, vertical, range).
In the Car Park experiment, the radar scans contain approximately 5000 points.
The range of the sensor is over \SI{350}{\meter} and the manufacturer indicates that adaptive modulation is used to boost resolution while maintaining long range.
From the point clouds provided in the dataset, it is apparent that some enhancement is applied by the sensor software.
Zhang's sensor rig, shown in \FIG{fig:carpark_detail}, also includes a lidar, barometer, camera and two \acp{IMU}, a standalone Vectornav IMU and an internal \ac{IMU} of the lidar sensor.
For testing the odometry variants that require inertial data, we use the Vectornav \ac{IMU} measurements.
The trajectory of the Car Park experiment is a rectangle recorded by a hand-pushed trolley with the sensor rig attached to it.
The environment is a parking lot between buildings at a university campus.
The pre-computed ground-truth localization based on a lidar \ac{SLAM} solution is available in the dataset and used by us.

\begin{figure*} 
    \centering   
    \begin{subfigure}[b]{0.49\textwidth}
        \includegraphics[width=\textwidth, trim=35 0 55
        8, clip]{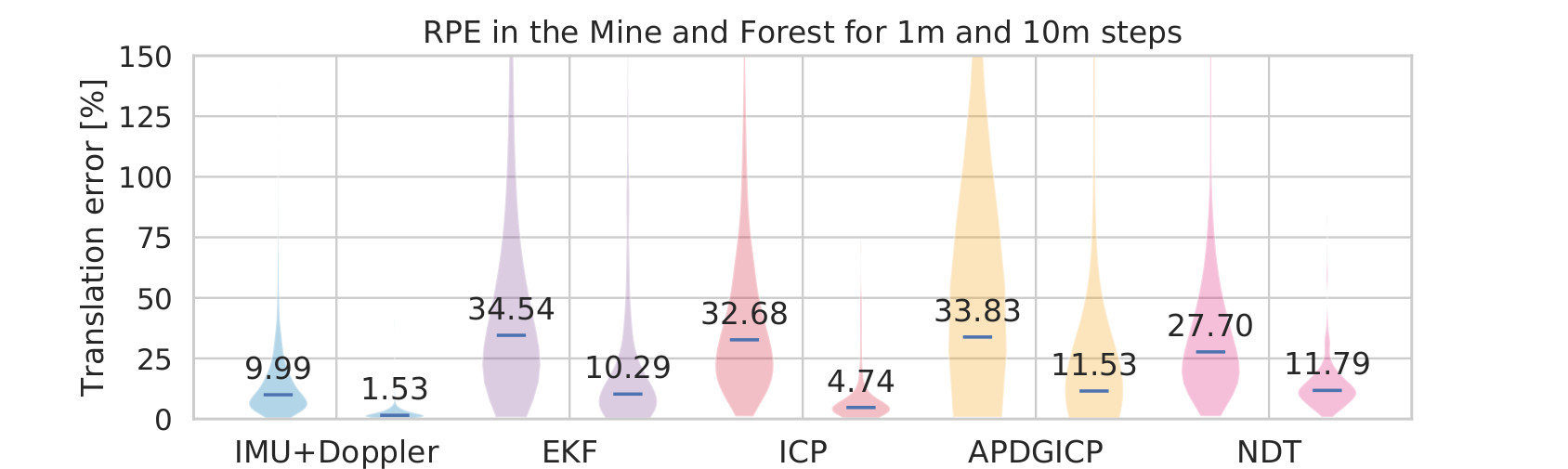}
        
        \vspace{10pt}
        
        \includegraphics[width=\textwidth, trim=35 0 55
        20, clip]{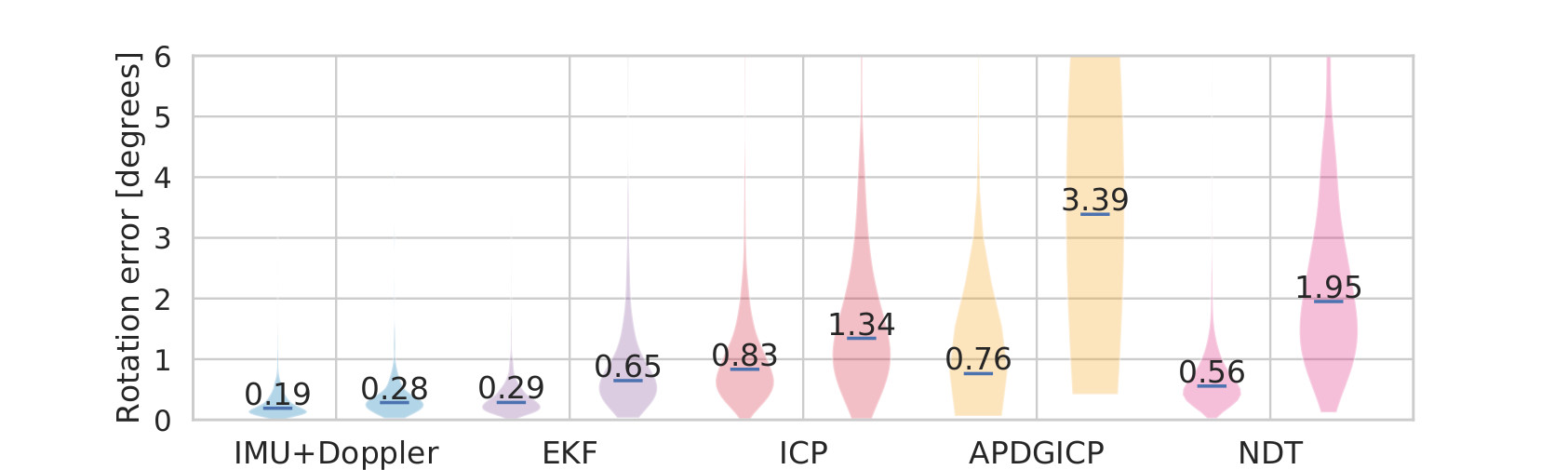}
        \label{subfig:mine_forest}
    \end{subfigure}
    \begin{subfigure}[b]{0.49\textwidth}
        \includegraphics[width=\textwidth, trim=35 0 55
        8, clip]{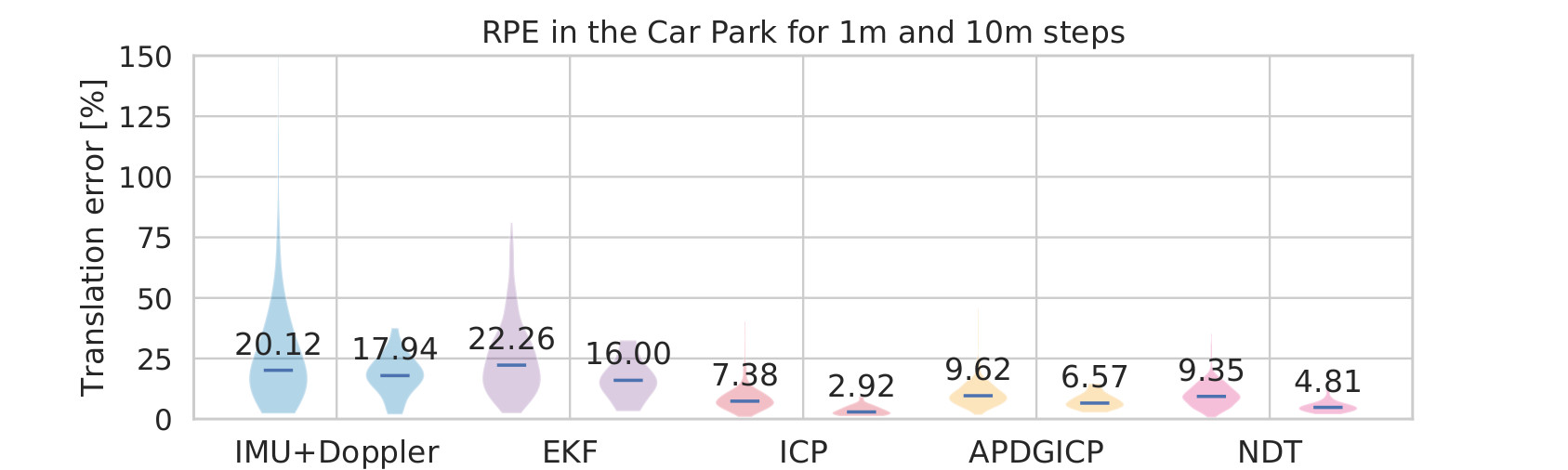}
        
        \vspace{10pt}
        
        \includegraphics[width=\textwidth, trim=35 0 55
        20, clip]{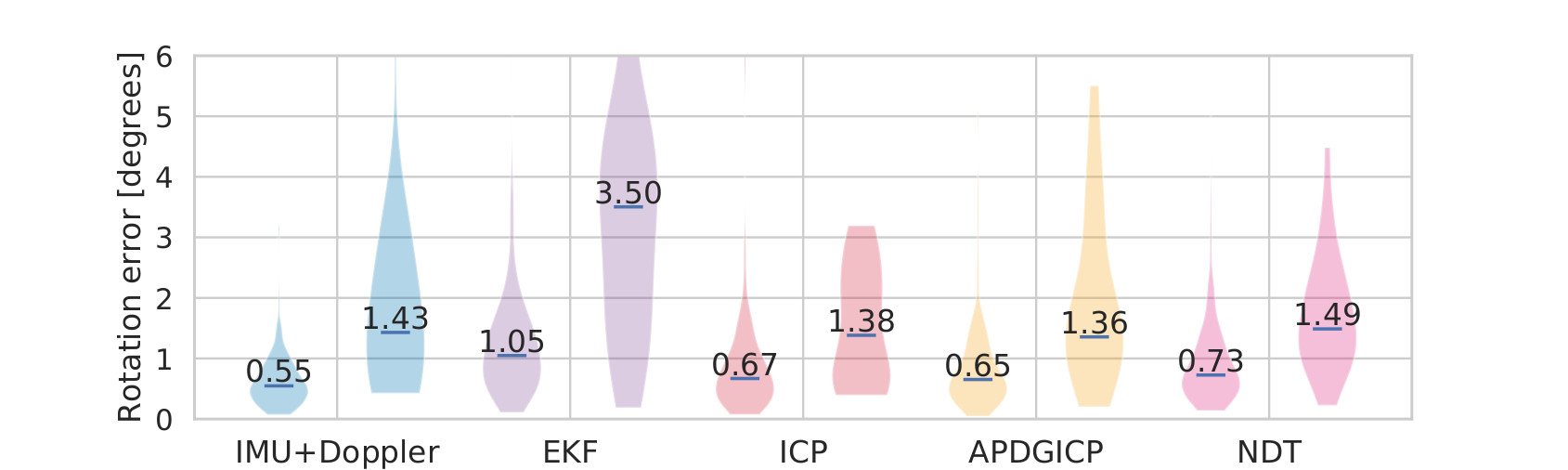}
        \label{subfig:cp}
    \end{subfigure}
    \caption{RPE values for the two distinct sensor setups. Each pair of the violin plots represents the step sizes for evaluating RPE, $1$m and $10$m respectively. The median RPE value is shown directly in each plot.}
    \label{fig:rpes}
\end{figure*}

\subsection{Odometry performance evaluation}
\label{sec:odom_performance}

To compare the performance of the radar odometry variants presented in \SEC{sec:odom_variants}, we use the widely adopted \ac{APE} and \ac{RPE} metrics, in the \emph{Evo} library implementation ~\cite{grupp2017evo}.
\ac{APE} together with trajectory plots provides the initial, general idea about the odometry variant behavior for the given sensory and environmental combinations, but is susceptible to the multiplicative, nonlinear effect of the accumulated attitude error.
\ac{RPE} complements this metric, with the indication of the rate of error accumulation.
For \ac{APE}, we provide its translation component, as the rotation error is apparent in the accompanying trajectory plots.
For \ac{RPE}, we provide an overall statistic for both the translation and rotation component.
The ground truth for evaluating the odometry error
comes from lidar-based \ac{SLAM}.
The lidar map and subsequently the reference localization was created by the open-source \emph{HDL Graph Slam}~\cite{Koide2019} implementation. 

\FIG{fig:trajectory_mine} and \FIG{fig:ape_kvarn} demonstrate the perfomance of the discussed radar odometries in the \textbf{Mine} experiment.
The scan-to-scan matching \textbf{APDGICP} variants (with and without our prior provided) together with NDT are not suitable for the type of output the Hugin radar provides.
We omit them from the \FIG{fig:trajectory_mine} plot since they randomly diverge, as can be seen in the \ac{APE} plot.
We attribute this to the low density and high variance in subsequent radar scans, which causes the scan-to-scan matching approach to quickly diverge.
This behavior is similar also in the case when we provide the more accurate IMU+Doppler prior estimate.
The main source of error, as we later show in \ac{RPE}, is the strong drift in attitude.

The scan-to-submap matching represented by the \textbf{ICP} variants (4DOF and 6DOF) performs better in the Mine experiment, although the drift is much stronger compared to what would be expected with lidar odometry in similar environments (e.g., refer to the SLAM results of the Subterranean DARPA robotic challenge \cite{ebadi2022present}).
Constraining the ICP to 4DOF reduces the vertical drift and results in overall lower \ac{APE}.

Comparable results are obtained from the \textbf{EKF} approach, which is free from the scan-matching problems, but suffers from abrupt changes in the measured Doppler velocity.
As long as the ground is smooth, the localization drift is comparable to lidar odometry drift rates, \ie below 1\%.
Once the truck hits a bump, the EKF reacts by inappropriate corrections, which can be observed at 180 and 250 seconds in the Mine experiment.
As the Hugin radar does not provide a measurement trigger signal, we assume that the measurement lag causes mismatch with the inertial measurements.
The radar scans are time-stamped, however the available EKF implementation does not recompute the past states, it rather counts on timely trigger signals and the \emph{state cloning} technique.
This problem can be partially alleviated by increasing the measurement uncertainty in the Doppler velocity, which makes the estimated trajectory smoother, but also reduces the EKF capacity to quickly estimate sensor biases and to sense minor motions.
We denote this altered variant as \emph{Doppler dampened} in the plots.

Surprisingly, the simplest \textbf{IMU+Doppler} approach shows the best results.
The drift is minimal, comparable to the best state-of-the-art lidar odometry techniques.
We attribute this to the high accuracy of the Hugin radar in the Doppler velocity values, and to the capability of the particular \ac{IMU} unit to suppress the heading drift by benefiting from the magnetometer measurements.
The downside is that it does not, contrary to the other techniques, provide any confidence estimate.

The results from the \textbf{Forest} experiment follow the trend of the Mine experiment.
\FIG{fig:ape_forest} shows that the scan-to-scan techniques diverge immediately and the scan-to-submap ICP drifts with similar rate compared to the Mine experiment.
The main difference is in the behavior of the Doppler dampened EKF. 
Thanks to the slower pace and the overall stability of the large and heavy wheel loader, it does not suffer from the abrupt Doppler velocity jolts and closely follows the simple IMU+Doppler odometry.

In the \textbf{Car Park} experiment from the dataset based on the Oculii Eagle radar, we see a different trend.
\FIG{fig:trajectory_cp} shows that the simpler methods, \ie \textbf{IMU+Doppler} and \textbf{EKF}, suffer from vertical and heading drift.
We assume that this is mainly due to the type of \ac{IMU} used by \cite{Zhang2023} when recording the dataset.
The Doppler dampened EKF is omitted from the trajectory plot because it immediately diverges due to accelerometer bias, which takes a minute to estimate (see \FIG{fig:ape_cp}). 
Moreover, the Doppler velocity estimation is less accurate in this dataset, which affects the smoothness of the trajectory.
We assume that the scan enhancement process inside the sensor may affect the quality of the Doppler velocity values.
On the other hand, all the scan-matching techniques perform well.
The longer range and the adaptive modulation in the Eagle radar make the the task of scan matching more reliable.
In fact, all variants of ICP, APDGICP and NDT perform similarly and stay within \SI{10}{\meter} in APE as shown in \FIG{fig:ape_cp}.

We summarize the performance of the odometry methods with the two distinct radars in \FIG{fig:rpes} using the \ac{RPE} metric.
For clarity, we omit the sub-variants in this plot as their \ac{RPE} does not differ substantially.
The trajectories are divided into \SI{1}{\meter} and \SI{10}{\meter} steps for the \ac{RPE} evaluation.
The plot shows the distribution of the translation and rotation errors with the median value directly in the plot.
The translation error is expressed as percentage of the step, the rotation error is left in the absolute value, therefore the longer steps yield larger rotational error.
Also note that in translation, we observe higher relative errors in the \SI{1}{\meter} steps.
Given the already high accuracy of all the methods, we are approaching the noise in the ground-truth localization based on the lidar \ac{SLAM}.
This is also why we do not consider single-frame-sized steps, for which more accurate reference would be necessary.

The \textbf{Mine} and \textbf{Forest} experiments with the Hugin radar are mainly affected by the noise in the estimated attitude, as the translation error differences between the methods are not as profound as the resulting \ac{APE}.
The rotation part of the \ac{RPE} metric confirms the trend seen in \ac{APE}.
The raw orientation provided by the Xsens IMU supports the highly accurate Doppler velocity and leads to the highly accurate results.

The \textbf{Car Park} experiment reveals that the translation is worse for the methods depending on the Doppler velocity (IMU+Doppler, EKF).
In the rotation errors, we see the limiting effect of the scan matching, which prevents larger errors to accumulate, contrary to IMU+Doppler and EKF.

\section{CONCLUSIONS}

In this work, we have compared several radar odometry estimation methods on three datasets recorded in underground and outdoor environments with two distinct modern imaging radars.
With the Oculii Eagle radar, the scan-matching methods achieved higher accuracy than the filtering methods.
On the other hand, thanks to the highly accurate Doppler velocity measurement in the Sensrad Hugin radar, the simplest sensor fusion method IMU+Doppler achieves only 0.3\% position drift in the Mine and Forest experiments.
This makes the method suitable for resource-constrained machines  operating in harsh environments, such as heavy machinery in the mining industry.

In future work, we will investigate the source of the inaccuracy of the Doppler velocity in the Eagle radar and extend the radar odometry into a full \ac{SLAM} solution.
 
\newpage
 
\bibliographystyle{IEEEtran}
\bibliography{IEEEabrv,bibtex/biblio}

\end{document}